\documentclass[11pt]{article}

\usepackage{naacl2021}

\usepackage{times}
\usepackage{latexsym}

\usepackage[T1]{fontenc}
\usepackage[utf8]{inputenc}

\usepackage{amsfonts,amssymb,amsmath}
\usepackage{tabu}
\usepackage{algorithm}
\usepackage{algorithmic}
\usepackage{xspace}
\usepackage{subfigure}
\usepackage{graphicx}
\usepackage{paralist}
\usepackage[switch]{lineno} 
\usepackage{paralist}
\newcommand{\method}{\textsc{Cunmt}\xspace}
\newcommand{\en}{\texttt{En}\xspace}
\newcommand{\fr}{\texttt{Fr}\xspace}
\newcommand{\de}{\texttt{De}\xspace}
\newcommand{\ro}{\texttt{Ro}\xspace}
\newcommand{\zh}{\texttt{Zh}\xspace}

\newcommand{\ml}{\mathcal{L}}
\newcommand{\me}{\mathcal{E}}
\newcommand{\mw}{\mathcal{W}}

\newcommand{\lb}{\mathcal{B}}

\title{Cross-lingual Supervision Improves Unsupervised Neural Machine Translation}

\author{Mingxuan Wang$^1$ Hongxiao Bai$^2$
          Hai Zhao$^2$  Lei Li$^1$\\
  $^1$ByteDance AI Lab, Beijing, China\\
        { \{wangmingxuan.89,lileilab\}@bytedance.com} \\  $^2$ Department of ComputeScience and Engineering, Shanghai Jiao Tong University \\  \{baippa, zhaohai\} @cs.sjtu.edu.cn \\ }

 \date{}

\begin{document}
%\linenumbers  %
\maketitle

\begin{abstract}
    %Neural machine translation~(NMT) already achieves impressive success for language pairs with large amounts of parallel corpus. 
%It is challenging to train neural machine translation (NMT) without parallel data. Can one utilize cross-lingual supervision such as \en-\fr to elevate a specific unsupervised direction such as \en-\de is still an open problem.  
%Inspired by the success of transfer learning in low-resourse NMT,
We propose to improve unsupervised neural machine translation with cross-lingual supervision (\method),
which utilizes  supervision signals from high resource language pairs to improve the translation of zero-source languages.
Specifically, for  training \texttt{En-Ro} system without parallel  corpus, we can leverage the corpus from  \texttt{En-Fr} and \texttt{En-De} to collectively train the translation from one language into many languages under one model.
% \method is based on multilingual models which require no changes to the standard unsupervised NMT. 
Simple and effective,  \method significantly improves the translation quality with a big margin in the benchmark unsupervised translation tasks, and even achieves comparable performance to supervised NMT. 
In particular, on WMT'14 \en-\fr tasks \method achieves 37.6 and 35.18 BLEU score, which is very close to the large scale supervised setting and on WMT'16 \en-\ro tasks \method achieves 35.09 BLEU score which is even better than the supervised Transformer baseline. 

%In this work, we try to extend unsupervised machine translation to a multilingual setting and augment it with parallel data of rich resource languages.
%That is, we combine supervised and unsupervised neural machine translation into a multilingual machine translation system, which can achieve better results for some language pairs.
%By applying this settings, we can get a triangular training structure.
%Empirical results show that our method greatly improves the translation qualities on unsupervised pairs, and is beneficial to the low resource or zero-shot language translations.
    
\end{abstract}

\section{Introduction}
\label{sec:intro}
Neural machine translation~(NMT) has achieved great success and reached satisfactory translation performance for several language pairs~\cite{bahdanau_iclr2018_nmt,gehring2017cnnmt,vaswani2017transformer}.
Such breakthroughs heavily depend on the availability of colossal amounts of bilingual sentence pairs, such as the some 40 million parallel sentence pairs used in the training of  WMT14 English French Task.
As bilingual sentence pairs are costly to collect, the success of NMT has not been fully duplicated in the vast majority of language pairs, especially for zero-resource languages. 
Recently, \cite{artetxe2018unsupervisedmt,lample2018unsupervisedmt,} tackled this challenge by training unsupervised neural machine translation (UNMT) models using only monolingual data, which achieves considerably  high accuracy, but still not on par with that of the state of the art supervised models.

\begin{figure}[!ht] \centering
\subfigure[Pivot NMT]{
  \includegraphics[width=.45\linewidth]{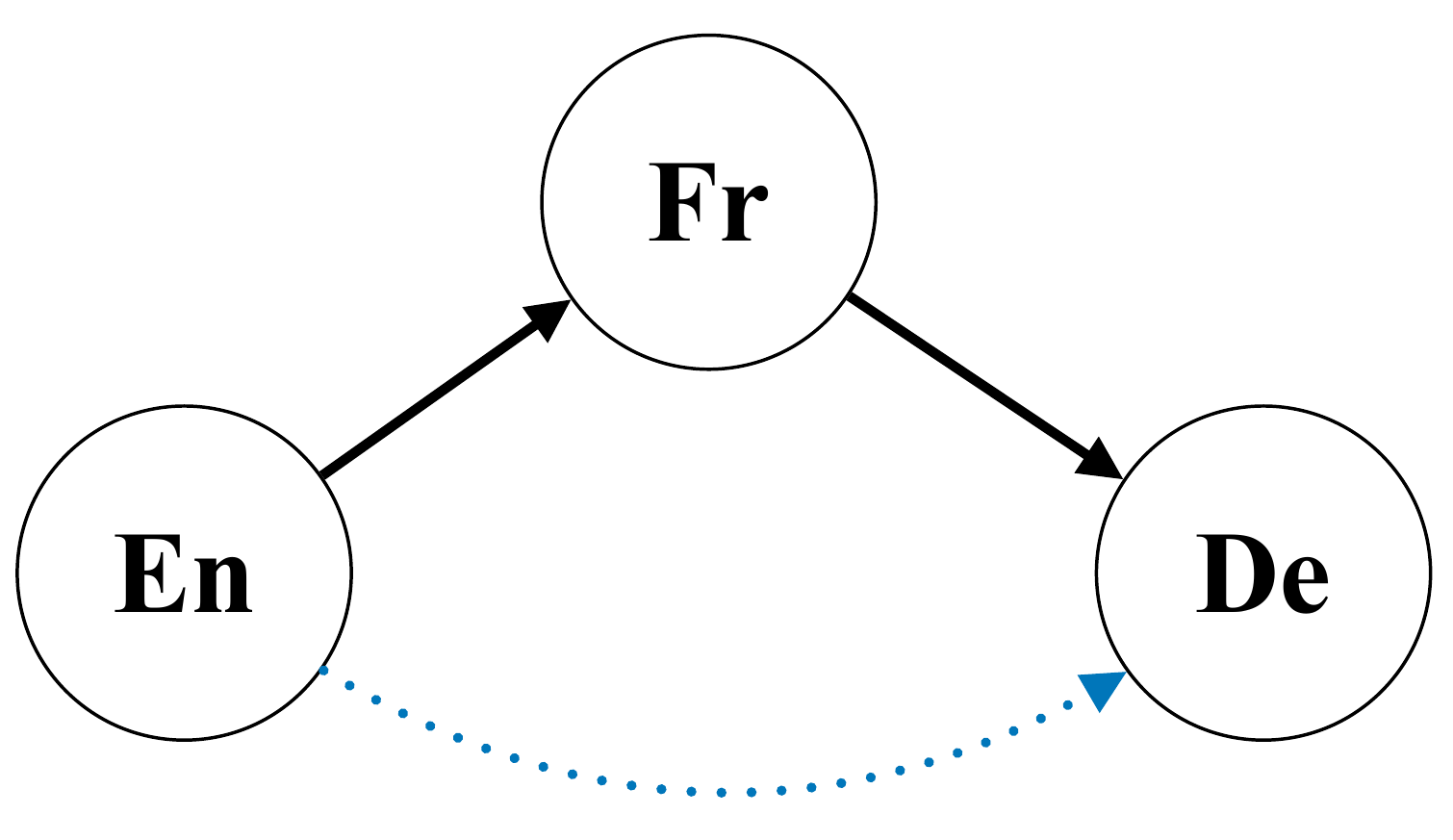}  
  \label{fig:struct1}
  }
  \subfigure[Unsupervised  NMT]{
  \includegraphics[width=.45\linewidth]{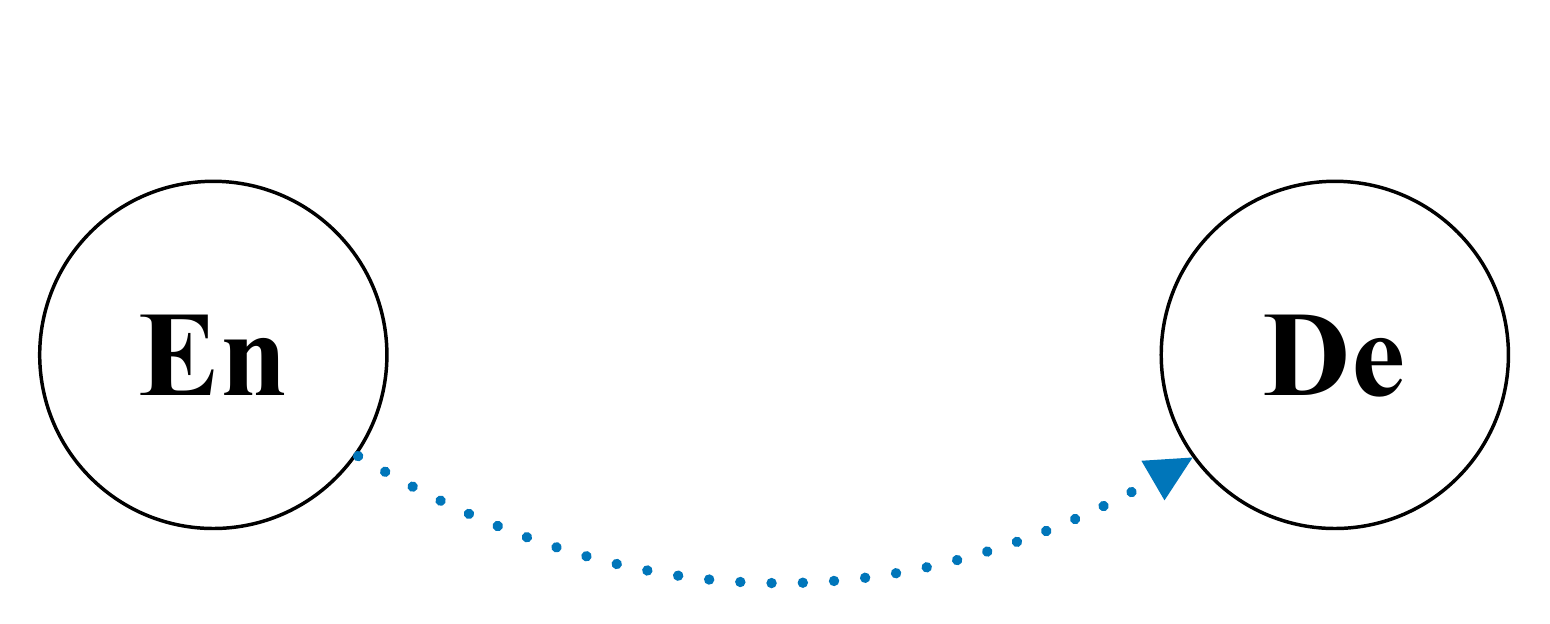}  
   \label{fig:struct2}
  }
  \subfigure[ \method w/o Para.]{
  \includegraphics[width=.45\linewidth]{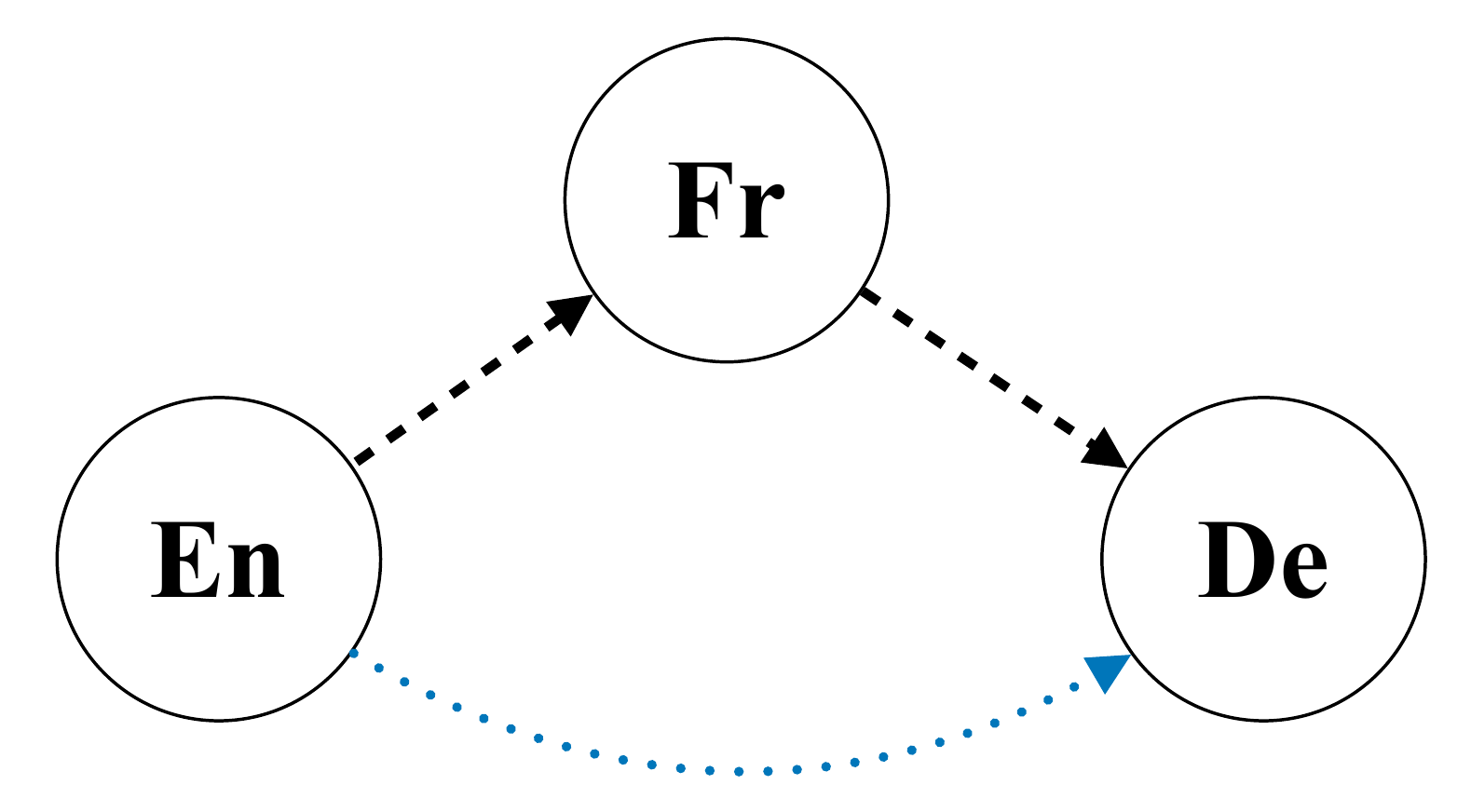}  
   \label{fig:struct3}
  }
  \subfigure[\method w/ Para. ]{
  \includegraphics[width=.45\linewidth]{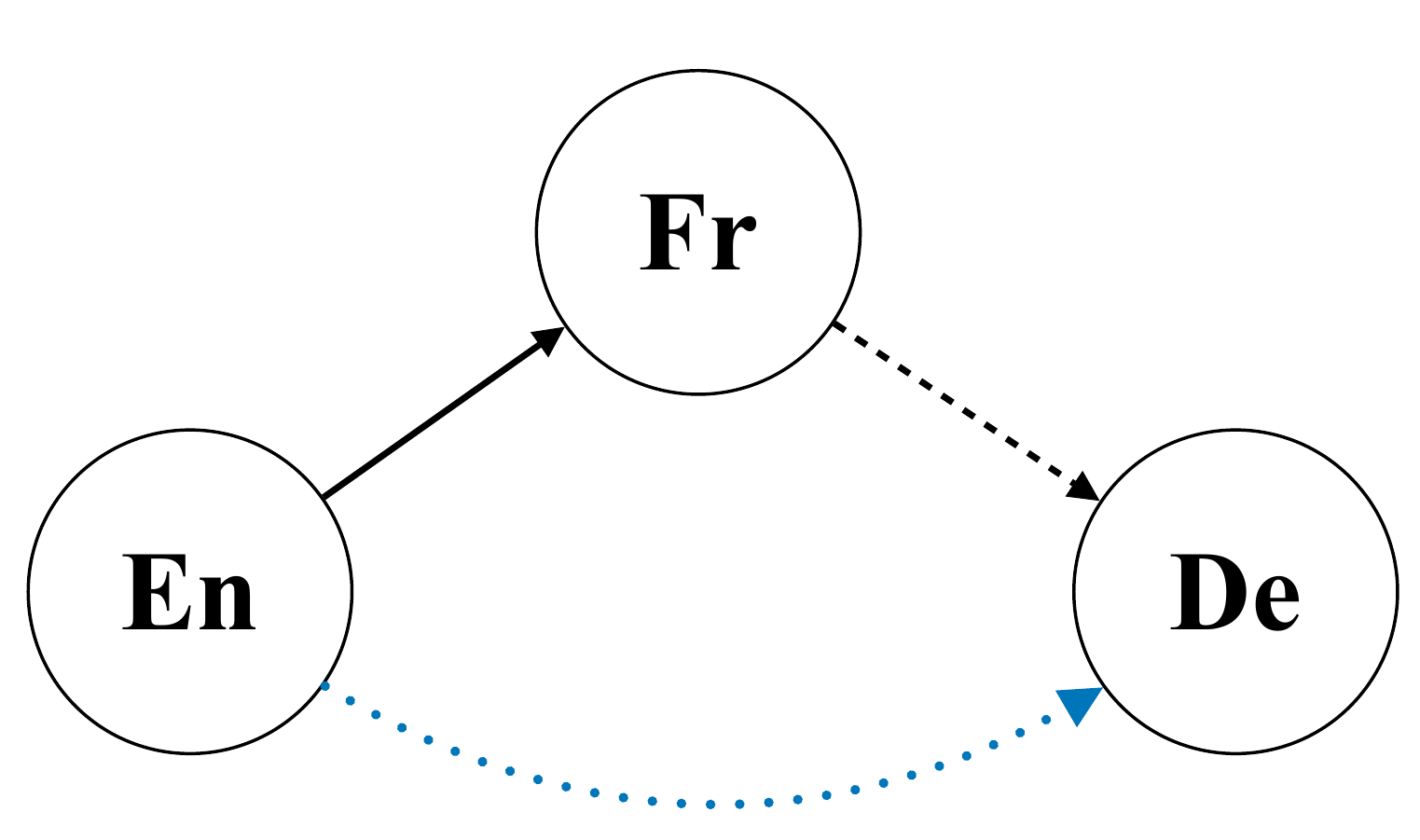}  
   \label{fig:struct4}
  }

\caption{Different settings for zero-resource NMT. Full edges  indicate the existence of parallel training data.  Dashed blue edges indicate the target translation pair.
``\method w/o Para.'' jointly train several unsupervised pairs in one model with unsupervised cross-lingual supervision.
``\method w/ Para.''  train unsupervised directions  with supervised cross-lingual supervision, such as jointly train unsupervised \texttt{En-De} with supervised \texttt{En-Fr}.}
\label{fig:struct}
\end{figure}

Most previous works focused on modeling the architecture  through parameter sharing or proper initialization to improve UNMT.
We argue that the  drawback of UNMT mainly stems from the lack of supervised signals, and  it is beneficial to transfer multilingual information across languages. 
In this paper, we take a step towards practical unsupervised NMT with cross-lingual supervision (\method) ---  making the most of the signal from other language. 
We  investigate  two variants of multilingual supervision for UNMT.  
\begin{inparaenum}[\it a)]
\item \method w/o Para.: a  general setting where  unrelated monolingual data can be introduced.  For example, using monolingual \fr data to help the training of \en-\de (Figure~\ref{fig:struct3}).
%Different from the previous zero-shot or pivot approaches, \method makes less assumptions and has a broader application scenario.
%To make the most of knowledge transformation from supervised signals, we then introduce a deliberately designed asymptotic fine-tuning method.
%Specifically,  
\item  \method w/ Para.: a relatively strict setting where other bilingual language pairs can be introduced. For example, we can naturally leverage parallel \en-\fr data to facilitate the unsupervised \en-\de translation (Figure~\ref{fig:struct4}). 
\end{inparaenum}

%Specifically, the cross-lingual signal is shared in two aspects. 
%First   different language pairs are trained  together, which enables transfer learning through parameter sharing. The idea has been proved to  be very effective  in low-resource machine translation. 
%When considering only monolingual  data, unrestricted
%\method takes advantage of multiple unsupervised translation tasks and jointly trains a singe model to serve all directions.
%This method  has been proven to be effective for supervised  NMT under low-resource scenario, but still lacks verification in UNMT.
%The method enables transfer learning through parameter sharing, however it still not effective enough and lack of guidance from supervision signal. 
%For the strict  scenario,  we  leverage the parallel data from other languages in two ways, shown in Figure~\ref{fig:struct3} and Figure~\ref{fig:struct4} respectively.
%Second, 
We introduce cross-lingual supervision which aims at modeling explicit translation probabilities across languages.
Taking three languages as an example, suppose the target  unsupervised direction is $\en \rightarrow \de$ and the auxiliary language is \fr.  Our target is to model the translation probability $p(\de|\en)$ with the support of $p(\fr|\en)$ and $p(\de|\fr)$. 
For  forward cross-lingual supervision,
the  system $\text{NMT}_{\fr \rightarrow \de}$ serves as a teacher, translating the  \fr part  of  parallel data $(\en, \fr)$ to  \de. 
The resulted synthetic data $(\en, \fr, \de)$ can be used to improve  our target system $\text{NMT}_{\en\rightarrow\de}$.
%Since $(\en, \fr)$ is supervised language pairs,  directly translating the sentence from \fr to \de to the error propagation.
For   backward cross-lingual supervision, we translate the monolingual \de to \fr with $\text{NMT}_{\de \rightarrow \fr}$, and then translate  \fr to \en with $\text{NMT}_{\fr \rightarrow \en}$. The resulted  synthetic bilingual data $(\de, \en)$ can be used for $\text{NMT}_{\en\rightarrow\de}$ as well. 

Our contributions can be summarized as follow:
\begin{inparaenum}[a)]
   % \item We propose to improve  UNMT with cross-lingual supervision, both explicit and implicit . The  proposed method is  capable of translating all  language pairs including both rich-resource and zero-resource ones all in one model , and making less assumptions.
    %\item  The utilization unrelated bilingual data and indirect supervision to assist the training of multilingual UNMT that was carried out in this study can be transferred to solve other unsupervised cross lingual problems.
    \item Empirical evaluation of \method on six  benchmarks  verifies that it surpassed individual MT models by a large margin of more than 3.0 BLEU points on average, and also bested several strong competitors. Particularly, on WMT'16 \en-\ro tasks, \method surpass the supervised baseline by 0.7 BLEU, showing  the great potential  for UNMT.
    \item  \method is very effective in the use of additional training data. MBART or MASS introduces billions of sentences, while \method only introduces tens of millions of sentences and achieves super or comparable  results. It shows the importance of introducing explicit supervision.
\end{inparaenum}

\section{The Proposed \method}
\label{sec:method}
\method  is based on a multilingual machine translation model involving supervised and unsupervised methods with a triangular training structure.  The original unsupervised NMT depends only on  monolingual corpus,  therefore the performances of these translation directions cannot be guaranteed.

Formally, given $n$ different languages $L_i$, $x_i$ denotes a sentence in language $L_i$.
$D_i$ denotes a monolingual dataset of $L_i$, and $D_{i,j}$ denotes a parallel dataset of $(L_i, L_j)$. We use $\mathcal{E}$ to indicate the set of all translation directions with parallel data and $\mathcal{W}$ to indicate the set of all unsupervised translation directions respectively. The goal of \method is to minimize the log likelihood of both unsupervised and supervised directions:
\begin{equation}
    \ml^\method = \sum_{i,j \in \mw }\ml^{U}_{i\to j} + \sum_{i,j \in \me} \ml^S_{i\to j} + \sum_{i,j \in \mw + \me }\hat \ml_{i\to j}
\end{equation}
where $\ml^{U}_{i\to j}$ is the unsupervised direct supervision, and $\ml^S_{i\to j}$ is the direct supervised supervision, and $\hat \ml_{i\to j} $ is the indirect supervision.
%The  settings of our method are mainly consistent with that of  \citet{lample2019cross}, but we only use pre-trained token embeddings for initialization rather than the pre-trained cross-lingual language model for time efficiency.

\subsection{Direct \& Cross-lingual Supervision}
\begin{figure} \centering
  \includegraphics[width=1\linewidth]{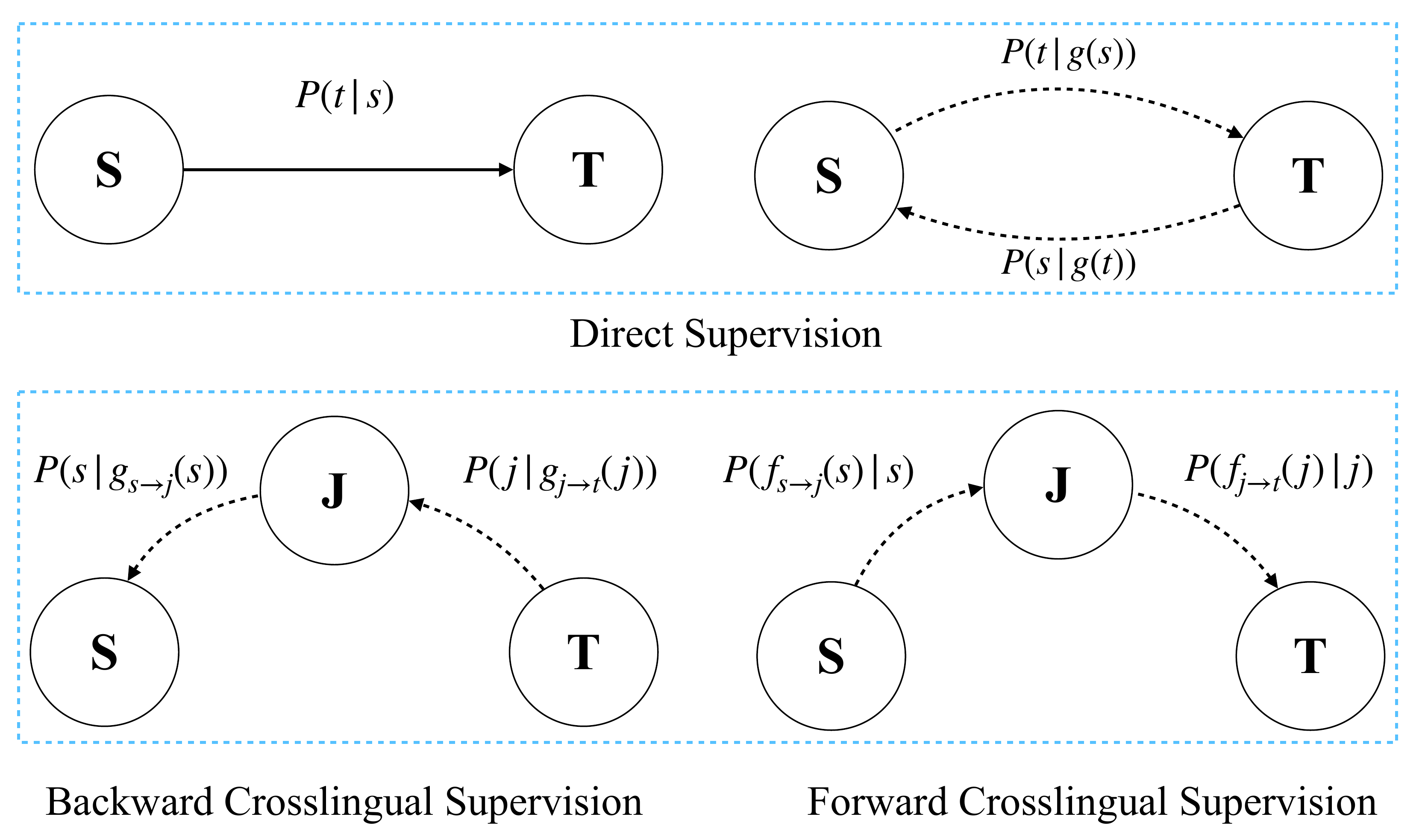}  
  \caption{Forward and backward cross lingual translation for auxiliary data. The dashed blue arrow indicates target unsupervised direction.  The solid arrow indicates using the parallel data. The dashed black arrow indicates generating synthetic data.}
  \label{fig:bt}
\end{figure}
\paragraph{Direct supervision} 
We will first introduce the notion of direct supervision loss, which only consider the translation probability between two different languages.

For supervised machine translation models, given parallel dataset $D_{s,t}$ with source language $L_s$ and target language $L_t$, we use $\ml^{S}_{s\to t}$ to denote the supervised training loss from language $L_s$ to language $L_t$. The training loss for a single sentence can be defined as:
\begin{equation}
\ml^{S}_{s\to t} = \mathbb{E}_{(x_s,x_t)\sim D_{s,t}} [-\log P(x_t|x_s)].
\label{eq:mt}
\end{equation}

For unsupervised machine translation models, only monolingual dataset $D_s$ and $D_t$ are given. We use $\ml^{U}_{s\to t}$ to denote the unsupervised training loss from language $L_s$ to language $L_t$. 
We use $\mathcal{B}_{s\to t}$ to denote this back translation procedure.
After that, we can use these data to train the model with supervised method from $L_s$ to $L_t$.
The losses of the dual structural are:
\begin{equation}
\begin{split}
\ml^\lb_{t\to s} = &\mathbb{E}_{x_s\sim D_s} [-\log P(x_s|g_{s\to t}(x_s)], \\
\ml^\lb_{s\to t} = &\mathbb{E}_{x_t\sim D_t} [-\log P(x_t|g_{t \to s}(x_t)],
\end{split}
\label{eq:bt}
\end{equation}
where $g_{s \to t}(x_s)$ translate the sentence in language $L_s$ to $L_t$, that is, the back translation of $x_s$.
Then the total loss of an unsupervised machine translation is:
\begin{equation}
    \ml^U = \ml^\lb_{t\to s}+ \ml^\lb_{s\to t}.
\end{equation}

%The initialization and back-translation are the most important parts of unsupervised machine translation.

\paragraph{Cross-lingual  supervision} 
When extend to the multilingual scenario, it is natural to introduce indirect supervision across languages. 
Given $n$ different languages, for each language pair $(L_i, L_j)$, we can easily obtain the translation probability of $P(x_i|x_j)$ through the direct supervised model $\ml^S$ or $\ml^U$. We use $ \hat\ml_{s\to t}$ to indicate the indirect supervised loss, which can be defined as:
\begin{equation}
    \hat\ml_{s\to t} = \sum_{i=0, i\neq s,t}^{n} \lambda_i \hat\ml_{s\to i \to t} 
\end{equation}
where $\lambda$ is the coefficient. T

Due to the lack of triples data $(L_i, L_k, L_j)$, it is difficult to directly estimate the cross translation loss $\hat\ml_{s\to i \to t}$. 
We therefor propose the backward and forward indirect supervision to calculate the cross loss:
\begin{equation} \label{eq:indirect}
    \begin{split}
            \hat\ml_{s\to j \to t} &= 
             \mathbb{E}_{x_t\sim D_t} [-\log P(x_t|g_{t\to j \to   s}(x_t))]  \\
             &+ \mathbb{E}_{x_s\sim D_s} [-\log P(f_{s\to j \to t}(x_s)|x_s)] 
    \end{split}
\end{equation}
where   $g_{t\to j \to s}(x_t)$ is the indirect backward translation  which translate $x_t$ to language $L_s$ and 
$f_{s\to j \to s}(x_t)$ is the indirect forward translation  which translate $x_s$ to language $L_t$.

\subsection{Training Procedure of \method}

The procedure of \method includes two main steps: multi-lingual pre-training and iterative multi-lingual training. 

\paragraph{Multi-lingual Pre-training}
Due to the ill-posed nature, it is also important to find a good  initialization to associate the source side languages and the target side languages.
We propose a Multi-lingual Pre-training approach, which jointly train the unsupervised auto-encoder and supervised machine translation. 
Intuitively, the multi-lingual joint pre-training can take advantage of transfer learning and thus benefit the low resource languages.
Apart form the monolingual data, pre-training can also leverage the bilingual parallel data. 
We suggest the supervised data provides strong signal to optimize the network, which  also advantage the unrelated unsupervised NMT pre-training. For example, it is beneficial to use the supervised \en-\fr model to initialize the unsupervised \de-\fr model. 

\paragraph{Indirect Supervised Training}
The goal is to train a single system that minimize the jointly loss function of $\ml^\method$.

Generally, \method can be applied to a restrict unsupervised scenario where only monolingual are provided, and also can be extended to a unrestricted scenario where parallel data are introduced.
For the sake of simplicity, we  describe our method on three language pairs, which can be easily extended to more language pairs.
Suppose that the three languages are denoted as the triad $(\en, \fr, \de)$, and we have monolingual data for all the three languages and also bilingual data for \en-\fr. The target is to train an unsupervised \en$\rightarrow$\fr system. 
The detailed method is as follows:
\begin{enumerate}
    \setlength{\itemsep}{0pt}
    \setlength{\parsep}{0pt}
    \setlength{\parskip}{0pt}
    \item Sample batch of monolingual $x_\en, x_\fr, x_\de$ sentences from $D_\en,D_\fr,D_\de$
    \item Sample batch of parallel sentence from $D_{\en,\fr}$ to generate supervised data $\mathcal{S}$
    \item Back translate $x_\en, x_\fr, x_\de$ to generate pseudo data $\mathcal{B}$
    \item Indirect back translate $x_\en, x_\fr, x_\de$ to generate pseudo data $\mathcal{B}^{i}$
    \item Indirect forward translate $x_\en, x_\fr, x_\de$ to generate pseudo data $\mathcal{F}^{i}$
    \item Merge $\mathcal{B}$, $\mathcal{B}^{i}$, $\mathcal{F}^{i}$ and $\mathcal{S}$ to jointly train \method.
    \item Repeat 1-6 until convergence. 
\end{enumerate}

For indirect or direct supervision, we follow the Equation~(\ref{eq:indirect}), which will adopts one step forward translation if parallel data is provided.
Since we train all directions in one model, the pseudo data will include all directions. In this setting, it contains: $\en\leftrightarrow \fr$, $\en\leftrightarrow\de$, $\fr\leftrightarrow\de$ with both direct and indirect directions.

\section{Experiments}
\label{sec:experiment}

\begin{table*}[!htb]
\centering
\begin{tabular}{lcccccc}
\hline
 & \multicolumn{2}{c}{$(\fr, \en, \de)$} & \multicolumn{2}{c}{$(\de,\en,\fr)$} & \multicolumn{2}{c}{$(\ro, \en, \fr)$} \\
  & \en-\fr & \fr-\en & \en-\de & \de-\en & \en-\ro & \ro-\en \\ 
  \hline
  Supervised Transformer & 41.0 & - & 34.0 & 38.6 & 34.3 & 34.0 \\
  \hline
  \multicolumn{7}{l}{Comparison systems of UNMT}\\
  \hline
%UNMT~\cite{artetxe2018unsupervised} &15.1 & 15.6 & 6.8 & 10.1& - & - \\
UNMT~\cite{lample-2018-pbumt} & 25.1 & 24.2 & 17.2 & 21.0 & 21.2 & 19.4 \\
EMB~\cite{lample2019cross} & 29.4 & 29.4 & 21.3 & 27.3 & 27.5 & 26.6 \\
MLM~\cite{lample2019cross} & 33.4 & 33.3 & 26.4 & 34.3 & 33.3 & 31.8 \\ 
MASS~\cite{song2019icml-mass} &37.5& 34.9 & 28.3 & \textbf{35.2} & \textbf{35.2} & 33.1 \\
MBART~\cite{liu2020multilingual} & - & - & \textbf{29.8} &34.0& 35.0 & 30.5  \\
\hline
  \multicolumn{7}{l}{\method}\\
  \hline
\method w/o Para. & 32.90 & 31.93 & 23.03 & 31.01 & 33.23 & 32.34 \\
\method w/ Para. & 34.37 & 32.77 & 23.99 & 31.98 & 33.95 & 33.15 \\
\method + Forward   & 35.88 & 33.64 & 26.50 & 33.11 & 34.12 & 33.61 \\
\method + Backward + Forward  & \textbf{37.60} & \textbf{35.18} & 27.60 & 34.10 & 35.09 & \textbf{33.95} \\
\hline
\end{tabular}
\caption{Main results comparisons.  MASS  uses large scale  pre-training and back translation during fine-tuning.  MBART employ large scale multi-lingual pretraining with billions sentences.
The last four lines are the results of our method.}
\label{table:res}
\end{table*}

\subsection{Datasets and Settings}

We conduct experiments including $(\de,\en,\fr)$, $(\fr, \en, \de)$, and $(\ro, \en, \fr)$.
For monolingual data of English, French and German,  20 million sentences from  available WMT monolingual News Crawl datasets were randomly selected.
For Romanian monolingual data,  all of the available Romanian sentences from News Crawl dataset were used and and were supplemented  with WMT16 monolingual data to yield a total of in 2.9 million sentences.
For parallel data, we use the standard WMT 2014 English-French dataset consisting of about 36M sentence pairs, and the standard WMT 2014 English-German dataset consisting of about 4.5M sentence pairs. For analyses, we also introduce the standard WMT 2017 English-Chinese dataset consisting of 20M sentence pairs. Consist with previous work, we report results on newstest 2014 for English-French pair, and on newstest 2016 for English-German and English-Romanian.

In the experiments, \method is built upon Transformer models.
We use the Transformer with 6 layers, 1024 hidden units, 16 heads.
We train our models with the Adam optimizer, a linear warm-up and learning rates varying from $10^{-4}$ to $5\times10^{-4}$.
The model is trained on 8 NVIDIA V100 GPUs.
We implement all our models in PyTorch based on the code of \cite{lample2019cross}\footnote{\url{https://github.com/facebookresearch/XLM}}.
All the results are evaluated on BLEU score with Moses scripts, which is in consist with the previous studies.

\subsection{Main Results}
The main results of similar pairs are shown in Table~\ref{table:res}. We make comparison  with three strong unsupervised methods:
\begin{itemize}
    \setlength{\itemsep}{0pt}
    \setlength{\parsep}{0pt}
    \setlength{\parskip}{0pt}
    \item MLM~\cite{lample2019cross} uses large scale cross-lingual data to train the mask language model and then fine-tune on unsupervised NMT.
    \item  MASS~\cite{song2019icml-mass} is a sequence to sequence model pre-trained with billions of monolingual data. 
    \item MBART~\cite{liu2020multilingual} introduces tens of billions monolingual data to pre-train a deep Transformer model. 
\end{itemize}
\emph{\method is very efficient in the use of multi-lingual data.} While the pretrained language model is obtained through several hundred times larger monolingual or cross-lingual corpus, 
\method achieves superior or comparable results with much less cost.  
%It should be noted that these competitors leverages several hundred times monolingual data to enhance the performance, while \method only introduces additional data from different languages. %The results demonstrate the effectiveness of \method for using the indirect supervision across languages.

The model was  improved by using synthetic data of cross translation that is based on the jointly trained model.
The results of ``\method + Forward'' are from the model tuned by only 1 epoch with about 100K sentences.
This method is fast and the performances are surprisingly effective.
The ``\method + Forward + Backward'' denotes that, besides forward translation, we also use monolingual data and cross translate it to the source language.
This method yielded the best performance by outperforming the ``\method w/o Para.'' by more than 3 BLEU score on average. The improvements show great potential for introducing indirect cross lingual supervision for unsupervised NMT.

When compared with supervised approaches, \method shows very promising performance. For the large scale WMT14 \en-\fr tasks, the gap between \method and supervised baseline is closed to 3.4 BLEU score. And for the medium  WMT16 \en-\ro task, \method performs even better than the supervised approach.

\section{Analyses}

In this part, we conduct several studies on \method to better understand its setting.
\paragraph{Backward or Forward}
\begin{figure}[!ht]
\centering
\includegraphics[width=0.5\textwidth]{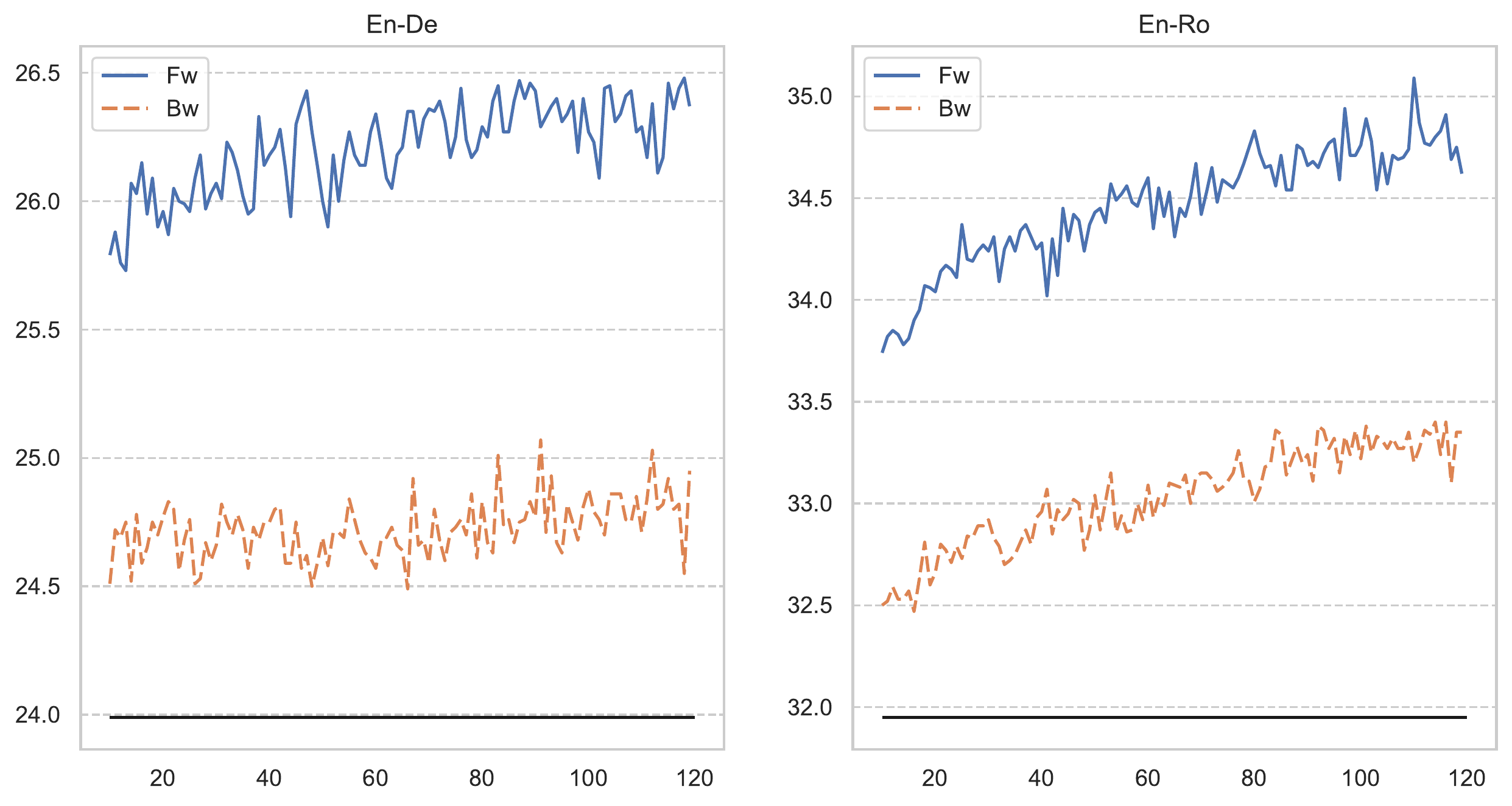}
\caption{Results comparison for \method fine-tuning with different auxiliary data.
%``Sv'' denotes fine-tuning with only \en-\fr parallel data. 
``Bw'' only adopts  cross-lingual backward  translation synthetic data, and ``Fw'' only adopts cross-lingual forward  translation synthetic data. 
The black horizontal is the baseline of UNMT.
The horizontal axis is epoch and the vertical axis is the BLEU score. Epoch size is 100K sentences.}
\label{fig:cur}
\end{figure}

 Here we have explored the effect of  cross-lingual backward supervision and cross-lingual forward supervision,  and plot the performance curves along with the training procedure in Figure~\ref{fig:cur}. 
 The comparison system is  \method trained only with monolingual data. To make a fair comparison, we use ``\method w/ Para.'' as the baseline model and fine-tuning it with only indirect forward supervision or indirect backward supervision. 
 We conduct experiments on WMT16 \en-\de and \en-\ro tasks. Clearly, the forward supervision outperforms the backward one with big margins, which shows the importance of introducing the forward supervision for multilingual UNMT. 
 It is still interesting to find that only introducing the indirect backward translation achieves better results than the unsupervised baseline. 
 
 We  suppose the reasons for the performance gap is that,
\begin{inparaenum}[\it a)]
    \item 
    The  UNMT baseline has included the traditional direct back translation, therefore the information gain from indirect backward  translation is limited compared to the forward translation. 
    \item The indirect forward translation provides a more direct way to model the relation  across different languages.  The results in consist with the previous research that pivot translation can help low resource language translation. 
\end{inparaenum}

\paragraph{Robustness on Parallel Data Scale}
\begin{table}[!htb]
\centering
\begin{tabu}{lcc}
\hline
  Auxiliary Direction &\en-\ro & \ro-\en \\ 
\hline
\en-\de & 34.86 & 33.18  \\
\en-\de $(50\%)$ & 34.72 & 32.85  \\ 
\en-\de $(25\%)$ & 34.52 & 32.33 \\ 
\hline
\end{tabu}
\caption{Robustness of Parallel Data Scale. Mainly evaluated on unsupervised \en-\ro direction with different auxiliary parallel data settings.}
\label{table:sd}
\end{table}
As shown in Table~\ref{table:sd},  \method is  robust to the parallel data scale. 
The results also dovetail with the unsupervised \en-\fr experiments in Table ~\ref{table:res}. 
As it turns out the smaller parallel data of \en-\de was able to significantly improve the performance of unsupervised \en-\fr translation. 
We then reduce the scale of the parallel data \en-\de and surprisingly find that even with only $25\%$ supervised data, \method still works well. The experiments demonstrate that \method is robust and has great potential to be applied to practical systems.

\paragraph{Importance of the Auxiliary Language}
\begin{table}[!htb]
\centering
\begin{tabu}{lcc}
\hline
  Auxiliary Direction &\en-\ro & \ro-\en \\ 
\hline
\en-\fr & 35.09 & 33.95  \\
\en-\de & 34.86 & 33.18  \\
\en-\zh & 33.85 & 32.86  \\
\en-\de-\fr & 35.26 & 34.20  \\
\hline
\end{tabu}
\caption{Effects of the Auxiliary Language. Mainly evaluated on unsupervised \en-\ro direction with different parallel data settings.\en-\fr,\en-\de and \en-\zh are the auxiliary parallel data for training \en-\ro.  \en-\de-\fr is the combination of the \en-\de and \en-\fr parallel data. }
\label{table:la}
\end{table}
Table ~\ref{table:la} shows effects of the auxiliary language.
We first switch the parallel data from \en-\fr to \en-\de, the performance is almost consistent. 
We then switch the parallen data  to $\en-\zh$, where \zh is dissimilar with $\ro$, the performance increases. This is in line with our expectations, that similar languages make it easier for transfer learning. 
Finally, we extend the parallel data to \en-\de and \en-\fr, and achieves further benefits.
Compared with \label{table:sd}, we suggest the language similarity is more important than the  auxiliary data scale.

\paragraph{Benefits as All in One Model}
\begin{table}[htb]
\centering
\begin{tabu}{lcc}
\hline
  System &\en-\fr & \fr-\en \\ 
\hline
Supervised Training & 39.70 & 36.62  \\
\method + Forward & 39.26 & 36.82  \\
\method + Backward & 39.12 & 36.20  \\ 
\hline
\end{tabu}
\caption{Translation performance on supervised directions of \method.}
\label{table:supres}
\end{table}

In table ~\ref{table:supres}, the performance of supervised directions are shown to illustrate the effects on which jointly training a single system has
First, we test the baseline supervised system, that is, only $En \to Fr$ and $Fr \to En$ are conducted on the model.
Due to difference in model architecture, the performance of \method is slightly lower than that of its state of the art counterparts. Also, some techniques such as model average are not applied, and two directions are trained in one model.
In \method, the performance of supervised directions drops a little, but in exchange, the performances of zero-shot directions are greatly improved and the model is convenient to serve for multiple translation directions.

\paragraph{Strategies of Synthetic Data Generation}
For the synthetic data generation, the reported results are from greedy decoding for time efficiency. 
We compared the effects of sample strategies 
on the language setting of $(\ro, \en, \de)$ where \en-\de is the supervised direction.
The results based on beam search generation for $\en \to \ro$ is 34.86, and 33.18 for $\en \to \fr$  in terms of BLEU.
Compared with greedy decoding, the performance of beam search  is  slightly inferior.
A possible reason is that the beam search makes the synthetic data further biased on the learned pattern. The results suggest that \method is exceedingly  robust to the sampling strategies when performing forward and backward cross translation. 

\section{Related Work}
\label{sec:related}
%Neural machine translation model relies heavily on large amounts of parallel data.
% However, it is difficult to obtain lots of parallel data for many language pairs.
%In order to better complete the low resource machine translation task, many efforts have been made to explore a way to improve system performance.
\paragraph{Multilingual NMT}
It has been proven low resource machine translation can adopt methods to utilize other rich resource data in order to develop a better system.
These methods include multilingual translation system~\cite{firat-etal-emnlp2016-mnmt,johnson-etal-2017-googlemnmt}, teacher-student framework~\cite{chen-etal-acl2017-teacher}, or others~\cite{zheng_ijcai2017-zeroresourcenmt}.
Apart from parallel data as an entry point, many attempts have been made to explore the usefulness of monolingual data, including semi-supervised methods and unsupervised methods which only monolingual data is used.
Much work also has been done to attempt to marry monolingual data with supervised data to create a better system, some of which include
using small amounts of parallel data and augment the system with monolingual data~\cite{sennrich-etal-acl2016-improvingwithmono,he_nips2016_dualnmt,wang_aaai2018_dualtransfernmt,gu-etal-naacl2018-lowresourcenmt,edunov-etal-emnlp2018-backtrans,yang2020towards}.
Others also try to utilize parallel data of rich resource language pairs and also monolingual data~\cite{ren-etal-acl2018-triangular,wang_iclr2018_multiagentnmt,al-shedivat-parikh-naacl2019-consistency,lin2020pre}.
\cite{ren-etal-acl2018-triangular} also proposed a triangular architecture, but their work still relied on parallel data of low resource language pairs.
With the joint support of parallel and monolingual data, the performance of a low resource system can be improved.

\paragraph{Unsupervised NMT}
In 2017, pure unsupervised machine translation method with only monolingual data was proven to be feasible.
On the basis of embedding alignment~\cite{artetxe-2017-unsupword,lample2018unsupword}, \cite{lample2018unsupervisedmt} and \cite{artetxe2018unsupervisedmt}  devised similar methods for fully unsupervised machine translation.
Considerable work has been done to improve the unsupervised machine translation systems by methods such as statistical machine translation~\cite{lample-2018-pbumt,artetxe-etal-emnlp2018-unsupervised,ren_aaai2019_usmt,artetxe-etal-acl2019-umt}, pretraining models~\cite{lample2019cross,song2019icml-mass}, or others~\cite{wu-etal-naacl2019-extracteditumt}, and all of which greatly improve the performance of unsupervised machine translation.

Our work attempts to utilize both monolingual and parallel data, and combine unsupervised and supervised machine translation through multilingual translation method into a single model \method to ensure better performance for unsupervised language pairs.

\section{Conclusion}
\label{sec:conclusion}

In this work, we propose a  multilingual machine translation framework \method incorporating distant supervision to tackle the challenge of the unsupervised translation task.
By mixing different training schemes into one model and utilizing  unrelated bilingual corpus, we greatly improve the performance of the unsupervised NMT direction.
By joint training, \method can serve all translation directions in one model. 
Empirically, \method has been proven to deliver substantial improvements over several strong UNMT competitors  and even achieve comparable performance to supervised NMT. 
In the future, we plan to build a universal \method system that is applicable in a wide span of languages.

% For future work, we will try to pre-train the cross-lingual model and build our method on the pre-trained model.

\bibliography{unmt}

\begin{thebibliography}{29}
\expandafter\ifx\csname natexlab\endcsname\relax\def\natexlab#1{#1}\fi

\bibitem[{Al-Shedivat and
  Parikh(2019)}]{al-shedivat-parikh-naacl2019-consistency}
Maruan Al-Shedivat and Ankur Parikh. 2019.
\newblock \href {https://doi.org/10.18653/v1/N19-1121} {Consistency by
  agreement in zero-shot neural machine translation}.
\newblock In \emph{Proceedings of the 2019 Conference of the North {A}merican
  Chapter of the Association for Computational Linguistics: Human Language
  Technologies (NAACL:HLT), Volume 1 (Long and Short Papers)}, pages
  1184--1197, Minneapolis, Minnesota.

\bibitem[{Artetxe et~al.(2017)Artetxe, Labaka, and
  Agirre}]{artetxe-2017-unsupword}
Mikel Artetxe, Gorka Labaka, and Eneko Agirre. 2017.
\newblock \href {https://doi.org/10.18653/v1/P17-1042} {Learning bilingual word
  embeddings with (almost) no bilingual data}.
\newblock In \emph{Proceedings of the 55th Annual Meeting of the Association
  for Computational Linguistics (ACL) (Volume 1: Long Papers)}, pages 451--462,
  Vancouver, Canada.

\bibitem[{Artetxe et~al.(2018{\natexlab{a}})Artetxe, Labaka, and
  Agirre}]{artetxe-etal-emnlp2018-unsupervised}
Mikel Artetxe, Gorka Labaka, and Eneko Agirre. 2018{\natexlab{a}}.
\newblock \href {https://doi.org/10.18653/v1/D18-1399} {Unsupervised
  statistical machine translation}.
\newblock In \emph{Proceedings of the 2018 Conference on Empirical Methods in
  Natural Language Processing (EMNLP)}, pages 3632--3642, Brussels, Belgium.

\bibitem[{Artetxe et~al.(2019)Artetxe, Labaka, and
  Agirre}]{artetxe-etal-acl2019-umt}
Mikel Artetxe, Gorka Labaka, and Eneko Agirre. 2019.
\newblock \href {https://www.aclweb.org/anthology/P19-1019} {An effective
  approach to unsupervised machine translation}.
\newblock In \emph{Proceedings of the 57th Annual Meeting of the Association
  for Computational Linguistics (ACL)}, pages 194--203, Florence, Italy.

\bibitem[{Artetxe et~al.(2018{\natexlab{b}})Artetxe, Labaka, Agirre, and
  Cho}]{artetxe2018unsupervisedmt}
Mikel Artetxe, Gorka Labaka, Eneko Agirre, and Kyunghyun Cho.
  2018{\natexlab{b}}.
\newblock \href {https://openreview.net/forum?id=Sy2ogebAW} {Unsupervised
  neural machine translation}.
\newblock In \emph{International Conference on Learning Representations
  (ICLR)}.

\bibitem[{Bahdanau et~al.(2015)Bahdanau, Cho, and
  Bengio}]{bahdanau_iclr2018_nmt}
Dzmitry Bahdanau, Kyunghyun Cho, and Yoshua Bengio. 2015.
\newblock Neural machine translation by jointly learning to align and
  translate.
\newblock In \emph{International Conference on Learning Representations
  (ICLR)}.

\bibitem[{Chen et~al.(2017)Chen, Liu, Cheng, and
  Li}]{chen-etal-acl2017-teacher}
Yun Chen, Yang Liu, Yong Cheng, and Victor~O.K. Li. 2017.
\newblock \href {https://doi.org/10.18653/v1/P17-1176} {A teacher-student
  framework for zero-resource neural machine translation}.
\newblock In \emph{Proceedings of the 55th Annual Meeting of the Association
  for Computational Linguistics (ACL) (Volume 1: Long Papers)}, pages
  1925--1935, Vancouver, Canada.

\bibitem[{Edunov et~al.(2018)Edunov, Ott, Auli, and
  Grangier}]{edunov-etal-emnlp2018-backtrans}
Sergey Edunov, Myle Ott, Michael Auli, and David Grangier. 2018.
\newblock \href {https://doi.org/10.18653/v1/D18-1045} {Understanding
  back-translation at scale}.
\newblock In \emph{Proceedings of the 2018 Conference on Empirical Methods in
  Natural Language Processing (EMNLP)}, pages 489--500, Brussels, Belgium.

\bibitem[{Firat et~al.(2016)Firat, Sankaran, Al-onaizan, Yarman~Vural, and
  Cho}]{firat-etal-emnlp2016-mnmt}
Orhan Firat, Baskaran Sankaran, Yaser Al-onaizan, Fatos~T. Yarman~Vural, and
  Kyunghyun Cho. 2016.
\newblock \href {https://doi.org/10.18653/v1/D16-1026} {Zero-resource
  translation with multi-lingual neural machine translation}.
\newblock In \emph{Proceedings of the 2016 Conference on Empirical Methods in
  Natural Language Processing (EMNLP)}, pages 268--277, Austin, Texas.

\bibitem[{Gehring et~al.(2017)Gehring, Auli, Grangier, Yarats, and
  Dauphin}]{gehring2017cnnmt}
Jonas Gehring, Michael Auli, David Grangier, Denis Yarats, and Yann~N. Dauphin.
  2017.
\newblock \href {http://proceedings.mlr.press/v70/gehring17a.html}
  {Convolutional sequence to sequence learning}.
\newblock In \emph{Proceedings of the 34th International Conference on Machine
  Learning (ICML)}, volume~70 of \emph{Proceedings of Machine Learning
  Research}, pages 1243--1252, International Convention Centre, Sydney,
  Australia.

\bibitem[{Gu et~al.(2018)Gu, Hassan, Devlin, and
  Li}]{gu-etal-naacl2018-lowresourcenmt}
Jiatao Gu, Hany Hassan, Jacob Devlin, and Victor~O.K. Li. 2018.
\newblock \href {https://doi.org/10.18653/v1/N18-1032} {Universal neural
  machine translation for extremely low resource languages}.
\newblock In \emph{Proceedings of the 2018 Conference of the North {A}merican
  Chapter of the Association for Computational Linguistics: Human Language
  Technologies (NAACL:HLT), Volume 1 (Long Papers)}, pages 344--354, New
  Orleans, Louisiana.

\bibitem[{He et~al.(2016)He, Xia, Qin, Wang, Yu, Liu, and
  Ma}]{he_nips2016_dualnmt}
Di~He, Yingce Xia, Tao Qin, Liwei Wang, Nenghai Yu, Tie-Yan Liu, and Wei-Ying
  Ma. 2016.
\newblock \href
  {http://papers.nips.cc/paper/6469-dual-learning-for-machine-translation.pdf}
  {Dual learning for machine translation}.
\newblock In \emph{Advances in Neural Information Processing Systems (NeurIPS)
  29}, pages 820--828.

\bibitem[{Johnson et~al.(2017)Johnson, Schuster, Le, Krikun, Wu, Chen, Thorat,
  Vi{\'e}gas, Wattenberg, Corrado, Hughes, and
  Dean}]{johnson-etal-2017-googlemnmt}
Melvin Johnson, Mike Schuster, Quoc~V. Le, Maxim Krikun, Yonghui Wu, Zhifeng
  Chen, Nikhil Thorat, Fernanda Vi{\'e}gas, Martin Wattenberg, Greg Corrado,
  Macduff Hughes, and Jeffrey Dean. 2017.
\newblock \href {https://doi.org/10.1162/tacl_a_00065} {{G}oogle{'}s
  multilingual neural machine translation system: Enabling zero-shot
  translation}.
\newblock \emph{Transactions of the Association for Computational Linguistics
  (TACL)}, 5:339--351.

\bibitem[{Lample and Conneau(2019)}]{lample2019cross}
Guillaume Lample and Alexis Conneau. 2019.
\newblock Cross-lingual language model pretraining.
\newblock \emph{arXiv preprint arXiv:1901.07291}.

\bibitem[{Lample et~al.(2018{\natexlab{a}})Lample, Conneau, Denoyer, and
  Ranzato}]{lample2018unsupervisedmt}
Guillaume Lample, Alexis Conneau, Ludovic Denoyer, and Marc'Aurelio Ranzato.
  2018{\natexlab{a}}.
\newblock \href {https://openreview.net/forum?id=rkYTTf-AZ} {Unsupervised
  machine translation using monolingual corpora only}.
\newblock In \emph{International Conference on Learning Representations
  (ICLR)}.

\bibitem[{Lample et~al.(2018{\natexlab{b}})Lample, Conneau, Ranzato, Denoyer,
  and Jégou}]{lample2018unsupword}
Guillaume Lample, Alexis Conneau, Marc'Aurelio Ranzato, Ludovic Denoyer, and
  Hervé Jégou. 2018{\natexlab{b}}.
\newblock \href {https://openreview.net/forum?id=H196sainb} {Word translation
  without parallel data}.
\newblock In \emph{International Conference on Learning Representations
  (ICLR)}.

\bibitem[{Lample et~al.(2018{\natexlab{c}})Lample, Ott, Conneau, Denoyer, and
  Ranzato}]{lample-2018-pbumt}
Guillaume Lample, Myle Ott, Alexis Conneau, Ludovic Denoyer, and Marc{'}Aurelio
  Ranzato. 2018{\natexlab{c}}.
\newblock \href {https://doi.org/10.18653/v1/D18-1549} {Phrase-based {\&}
  neural unsupervised machine translation}.
\newblock In \emph{Proceedings of the 2018 Conference on Empirical Methods in
  Natural Language Processing (EMNLP)}, pages 5039--5049, Brussels, Belgium.

\bibitem[{Lin et~al.(2020)Lin, Pan, Wang, Qiu, Feng, Zhou, and Li}]{lin2020pre}
Zehui Lin, Xiao Pan, Mingxuan Wang, Xipeng Qiu, Jiangtao Feng, Hao Zhou, and
  Lei Li. 2020.
\newblock Pre-training multilingual neural machine translation by leveraging
  alignment information.
\newblock \emph{arXiv preprint arXiv:2010.03142}.

\bibitem[{Liu et~al.(2020)Liu, Gu, Goyal, Li, Edunov, Ghazvininejad, Lewis, and
  Zettlemoyer}]{liu2020multilingual}
Yinhan Liu, Jiatao Gu, Naman Goyal, Xian Li, Sergey Edunov, Marjan
  Ghazvininejad, Mike Lewis, and Luke Zettlemoyer. 2020.
\newblock Multilingual denoising pre-training for neural machine translation.
\newblock \emph{arXiv preprint arXiv:2001.08210}.

\bibitem[{Ren et~al.(2018)Ren, Chen, Liu, Li, Zhou, and
  Ma}]{ren-etal-acl2018-triangular}
Shuo Ren, Wenhu Chen, Shujie Liu, Mu~Li, Ming Zhou, and Shuai Ma. 2018.
\newblock \href {https://doi.org/10.18653/v1/P18-1006} {Triangular architecture
  for rare language translation}.
\newblock In \emph{Proceedings of the 56th Annual Meeting of the Association
  for Computational Linguistics (ACL) (Volume 1: Long Papers)}, pages 56--65,
  Melbourne, Australia.

\bibitem[{Ren et~al.(2019)Ren, Zhang, Liu, Zhou, and Ma}]{ren_aaai2019_usmt}
Shuo Ren, Zhirui Zhang, Shujie Liu, Ming Zhou, and Shuai Ma. 2019.
\newblock \href {https://doi.org/10.1609/aaai.v33i01.3301241} {Unsupervised
  neural machine translation with smt as posterior regularization}.
\newblock \emph{Proceedings of the AAAI Conference on Artificial Intelligence
  (AAAI)}, 33:241--248.

\bibitem[{Sennrich et~al.(2016)Sennrich, Haddow, and
  Birch}]{sennrich-etal-acl2016-improvingwithmono}
Rico Sennrich, Barry Haddow, and Alexandra Birch. 2016.
\newblock \href {https://doi.org/10.18653/v1/P16-1009} {Improving neural
  machine translation models with monolingual data}.
\newblock In \emph{Proceedings of the 54th Annual Meeting of the Association
  for Computational Linguistics (ACL) (Volume 1: Long Papers)}, pages 86--96,
  Berlin, Germany.

\bibitem[{Song et~al.(2019)Song, Tan, Qin, Lu, and Liu}]{song2019icml-mass}
Kaitao Song, Xu~Tan, Tao Qin, Jianfeng Lu, and Tie-Yan Liu. 2019.
\newblock \href {http://proceedings.mlr.press/v97/song19d.html} {{MASS}: Masked
  sequence to sequence pre-training for language generation}.
\newblock In \emph{Proceedings of the 36th International Conference on Machine
  Learning (ICML)}, volume~97 of \emph{Proceedings of Machine Learning
  Research}, pages 5926--5936, Long Beach, California, USA.

\bibitem[{Vaswani et~al.(2017)Vaswani, Shazeer, Parmar, Uszkoreit, Jones,
  Gomez, Kaiser, and Polosukhin}]{vaswani2017transformer}
Ashish Vaswani, Noam Shazeer, Niki Parmar, Jakob Uszkoreit, Llion Jones,
  Aidan~N Gomez, \L~ukasz Kaiser, and Illia Polosukhin. 2017.
\newblock \href
  {http://papers.nips.cc/paper/7181-attention-is-all-you-need.pdf} {Attention
  is all you need}.
\newblock In \emph{Advances in Neural Information Processing Systems (NeurIPS)
  30}, pages 5998--6008.

\bibitem[{Wang et~al.(2018)Wang, Xia, Zhao, Bian, Qin, Liu, and
  Liu}]{wang_aaai2018_dualtransfernmt}
Yijun Wang, Yingce Xia, Li~Zhao, Jiang Bian, Tao Qin, Guiquan Liu, and Tie-Yan
  Liu. 2018.
\newblock \href {https://aaai.org/ocs/index.php/AAAI/AAAI18/paper/view/17041}
  {Dual transfer learning for neural machine translation with marginal
  distribution regularization}.
\newblock In \emph{Proceedings of AAAI Conference on Artificial Intelligence
  (AAAI)}, pages 5553--5560, New Orleans, USA.

\bibitem[{Wang et~al.(2019)Wang, Xia, He, Tian, Qin, Zhai, and
  Liu}]{wang_iclr2018_multiagentnmt}
Yiren Wang, Yingce Xia, Tianyu He, Fei Tian, Tao Qin, ChengXiang Zhai, and
  Tie-Yan Liu. 2019.
\newblock \href {https://openreview.net/forum?id=HyGhN2A5tm} {Multi-agent dual
  learning}.
\newblock In \emph{International Conference on Learning Representations
  (ICLR)}.

\bibitem[{Wu et~al.(2019)Wu, Wang, and Wang}]{wu-etal-naacl2019-extracteditumt}
Jiawei Wu, Xin Wang, and William~Yang Wang. 2019.
\newblock \href {https://doi.org/10.18653/v1/N19-1120} {Extract and edit: An
  alternative to back-translation for unsupervised neural machine translation}.
\newblock In \emph{Proceedings of the 2019 Conference of the North {A}merican
  Chapter of the Association for Computational Linguistics: Human Language
  Technologies (NAACL:HLT), Volume 1 (Long and Short Papers)}, pages
  1173--1183, Minneapolis, Minnesota.

\bibitem[{Yang et~al.(2020)Yang, Wang, Zhou, Zhao, Zhang, Yu, and
  Li}]{yang2020towards}
Jiacheng Yang, Mingxuan Wang, Hao Zhou, Chengqi Zhao, Weinan Zhang, Yong Yu,
  and Lei Li. 2020.
\newblock Towards making the most of bert in neural machine translation.
\newblock In \emph{Proceedings of the AAAI Conference on Artificial
  Intelligence}, volume~34, pages 9378--9385.

\bibitem[{Zheng et~al.(2017)Zheng, Cheng, and
  Liu}]{zheng_ijcai2017-zeroresourcenmt}
Hao Zheng, Yong Cheng, and Yang Liu. 2017.
\newblock \href {https://doi.org/10.24963/ijcai.2017/594} {Maximum expected
  likelihood estimation for zero-resource neural machine translation}.
\newblock In \emph{Proceedings of the Twenty-Sixth International Joint
  Conference on Artificial Intelligence (IJCAI)}, pages 4251--4257.

\end{thebibliography}
\bibliographystyle{acl_natbib}

\end{document}